\documentclass[review]{elsarticle}
\graphicspath{ {./images/} }
\usepackage{hyperref}
\usepackage{float}
\usepackage{verbatim} 
\usepackage{apalike}
\usepackage{xcolor}
\restylefloat{figure}
\restylefloat{table}
\usepackage{appendix}
\usepackage{amsmath, amsthm, amsfonts, amssymb, gensymb}
\usepackage{balance}
\newcommand{\fs}{\text{.} }
\newcommand{\com}{\text{,} }

\newcommand{\R}{\mathbb R}

\newcommand{\bb}{\mathbf{b}}

\newcommand{\bx}{\mathbf{x}}
\newcommand{\bh}{\mathbf{h}}

\newcommand{\bW}{\mathbf{W}}

\newcommand{\TP}{\mathrm{TP}}
\newcommand{\TN}{\mathrm{TN}}
\newcommand{\FP}{\mathrm{FP}}
\newcommand{\FN}{\mathrm{FN}}
\newcommand{\PPV}{\mathrm{Precision}}
\newcommand{\TPR}{\mathrm{Sensitivity}}
\newcommand{\etal}{\textit{et al. }}

\journal{Expert Systems with Applications}

\biboptions{authoryear}

\begin{document}
\begin{frontmatter}

\title{RobIn: A Robust Interpretable Deep Network for Schizophrenia Diagnosis}

\author[Northumbria]{Daniel Organisciak}
\ead{daniel.organisciak@northumbria.ac.uk}
\author[Durham]{Hubert P. H. Shum}
\ead{hubert.shum@durham.ac.uk}
\author[Lagos]{Ephraim Nwoye}
\ead{enwoye@unilag.edu.ng}
\author[Northumbria]{Wai Lok Woo\corref{cor1}}
\ead{wailok.woo@northumbria.ac.uk}
\cortext[cor1]{Corresponding author}

\address[Northumbria]{Department of Computer and Information Sciences, Northumbria University, Newcastle, UK
            }

\address[Durham]{Department  of  Computer  Science,  Durham University, Durham, UK
            }
            
\address[Lagos]{Department  of  Biomedical  Engineering,  University of Lagos, Lagos, Nigeria
            }





\begin{abstract}
Schizophrenia is a severe mental health condition that requires a long and complicated diagnostic process. However, early diagnosis is vital to control symptoms.
Deep learning has recently become a popular way to analyse and interpret medical data. Past attempts to use deep learning for schizophrenia diagnosis from brain-imaging data have shown promise but suffer from a large training-application gap - it is difficult to apply lab research to the real world.
We propose to reduce this training-application gap by focusing on readily accessible data. We collect a data set of psychiatric observations of patients based on DSM-5 criteria. Because similar data is already recorded in all mental health clinics that diagnose schizophrenia using DSM-5, our method could be easily integrated into current processes as a tool to assist clinicians, whilst abiding by formal diagnostic criteria.
To facilitate real-world usage of our system, we show that it is interpretable and robust. 
Understanding how a machine learning tool reaches its diagnosis is essential to allow clinicians to trust that diagnosis.
To interpret the framework, we fuse two complementary attention mechanisms, `squeeze and excitation' and `self-attention', to determine global attribute importance and attribute interactivity, respectively. The model uses these importance scores to make decisions. This allows clinicians to understand how a diagnosis was reached, improving trust in the model.  
Because machine learning models often struggle to generalise to data from different sources, we perform experiments with augmented test data to evaluate the model's applicability to the real world. We find that our model is more robust to perturbations, and should therefore perform better in a clinical setting. It achieves 98\% accuracy with 10-fold cross-validation.
\end{abstract}

\begin{keyword}
Schizophrenia \sep Deep Learning \sep Self-Attention \sep Interpretability \sep Out-of-Distribution 
\end{keyword}

\end{frontmatter}

\section{Introduction}
According to the World Health Organisation, schizophrenia can cause a greater level of disability than any other physical or mental illness \citep{WHO2011}. 
Diagnosis and treatment must be conducted as early as possible to improve outcomes, so that it does not reach this level of severity \citep{Hafner2006}.
However, the symptoms of schizophrenia are similar to those caused by drug use, brain tumours, and other mental health problems like bipolar disorder.
This makes schizophrenia notoriously difficult to diagnose, with clinicians going through an exhaustive process to rule out other potential causes. A minimum of six months of observation is required before a schizophrenia diagnosis can be provided, according to DSM-5 criteria \citep{DSM5}.
Our model predicts the diagnosis after six months from one observation session with 98\% accuracy.


Machine learning is increasingly being applied for medical diagnosis in the real world. Identifying problems at an earlier stage helps to manage them before they develop into a serious condition.
Employing machine learning for disease diagnosis can also reduce a considerable number of required man-hours: machine learning systems can quickly process information and provide a recommendation. This would allow more patients to receive high-quality early-stage treatment.

To use machine learning to diagnose schizophrenia, we construct a data set that complies with current clinical assessment guidelines.
Clinicians currently must assess patients according to the Diagnostic and Statistical Manual of Mental Disorders (DSM-5) \citep{DSM5} criteria A-F. 
We compose a data set of clinical observations of patients with a mental health condition.
The collected attributes in the data set include attributes related to DSM-5 criterion A symptoms 
and attributes related to the symptoms of other mental health illnesses, to help the model to exclude these illnesses in the same way that psychiatrists must.
Moreover, mental health clinics that use DSM-5 criteria will record psychiatric observational data similar to ours. Our model can therefore be applied to data that is already recorded, or after a clinical appointment, with minimal change to current processes required. 
These data set design choices have been made to demonstrate the ability of our proposed framework to function in the real world. Many machine learning models have shown good performance on laboratory-created data sets, but their real-world impact has been minimal \citep{tandon2019using}. This paper aims to minimise the gap between research and practice.

Deep Neural Networks (DNNs) have become the most popular machine learning tool in recent years as they automatically learn effective features from the data,
rather than rely on human input to design hand-crafted features. Even with expert knowledge, these algorithms were sometimes ineffective, or had unwittingly incorporated the human experts' bias.
Therefore, DNNs typically attain stronger performance and generalisation ability compared with traditional machine learning algorithms \citep{alom2018history}. 

\begin{figure}[t]
\centering
\includegraphics[width=0.9\linewidth]{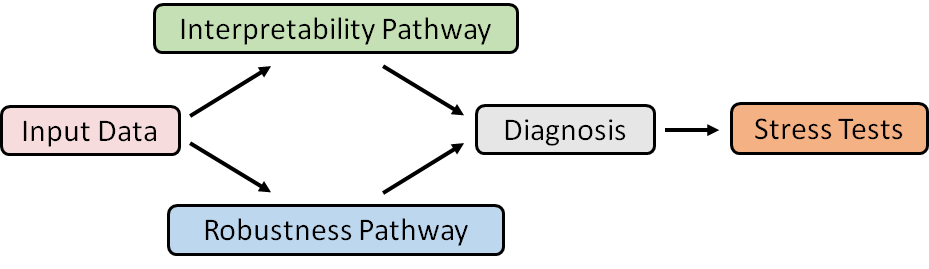}
\caption{An overview of our proposed method. Data is processed via two pathways: one for robustness and one for interpretability. These pathways are complementary to one another, provide an insight into how the model arrives at a decision, and can generalise to new distributions.} \label{fig:overview}
\end{figure}

One major drawback of DNNs is a lack of interpretability.
A missed diagnosis could result in death, whereas incorrectly diagnosing a patient with a condition could also result in complications (e.g. due to side-effects of needlessly prescribed medicine).
For a clinician to have confidence in a recommendation from a deep learning system, they must be able to understand why the system has come to that recommendation.

We propose a \emph{collaborative attention network} to improve the interpretability of the system and provide a high quality diagnosis in an efficient manner. It is important to consider \emph{static} and \emph{dynamic attention}. We wish to determine which features the network believes are important in general, and to understand how this changes based on specific values that features take, and combinations of those values. Our mechanism consists of a \emph{squeeze and excitation} (SE) module \citep{Hu_2018_CVPR} to compute the global feature importance, and a \emph{self-attention} (SA) module \citep{Vaswani2017} to decipher how features interact individually and with each other.

To deploy a machine learning model in a real-world setting, it must be robust when generalising to data from a different distribution.
A typical machine learning experiment will split a data set into a train and a test set. However, despite the test set being unseen during training, its distribution is often independent and identically distributed (i.i.d) to the training set. 
When a trained model is tested on distributions that are not identical to the training data, it often struggles to generalise to the new distribution \citep{d2020underspecification}.

Squeeze and Excitation networks \citep{Hu_2018_CVPR} compute very discriminative feature representations by learning which input features at each layer are most important. We posit that this fixed global feature importance allows the network to remain robust to data from new distributions. To show the effectiveness of our method, we design three stress tests at four different strength levels to alter the distribution of test data. Our stress tests consist of noise addition, data erasing and a combination of the two. We evaluate all compared models across these twelve robustness experiments to determine which would be suitable for deployment in a real-world clinical setting.

An overview of our paper can be found in Figure \ref{fig:overview}. We provide the following contributions: 
\begin{enumerate}
\item The collection of psychiatric evaluation data that enables machine learning training for automated diagnosis of schizophrenia.
\item A \textbf{Rob}ust, \textbf{In}terpretable (RobIn) deep network is developed to model the data to recommend a schizophrenia diagnosis. Our framework outperforms existing machine learning methods such as support vector machines, multi-layer perceptrons, and deep neural networks. 
\item RobIn is made interpretable on two levels: the adaptation of channel attention to determine global feature importance, and self-attention for feature interactivity analysis.
\item A total of twelve stress tests are designed to evaluate model performance on a distribution that is not i.i.d. to the training data, to simulate model performance in the real world. Our proposed model outperforms all other methods at remaining robust to perturbations.
\end{enumerate}

We perform experiments on two settings: the standard 90/10 cross-validation protocol that is most commonly used with small data sets, and a 50/50 train/test split with 25 runs as a baseline with which to compare the robustness stress tests. We find that RobIn outperforms other methods on both of these test settings, and is also most robust to distribution shift proposed to test robustness.

The rest of the paper is organised as follows. Related studies are outlined in Section \ref{sec:rw}. Section \ref{sec:data} provides more details on the data set. Section \ref{sec:DNN} gives details on the deep learning framework that we propose. Section \ref{sec:HUANN} outlines the attention mechanism helping clinicians to understand the outcomes of the DNN. Section \ref{sec:eval} contains our experimental evaluation and stress tests.
The paper is concluded in Section \ref{sec:conclusion}.

\section{Related Work}\label{sec:rw}   
\subsection{Diagnosis of Schizophrenia}
Schizophrenia has traditionally been diagnosed via specifications from the Diagnostic and Statistical Manual of Mental Disorders (DSM) or the International Classification of Diseases (ICD), with the most recent versions being DSM-5 \citep{DSM5} and ICD-11 \citep{Khoury2017} respectively. These traditional specifications recommend a diagnosis based on the symptoms that a patient is exhibiting.

The Research Domain Criteria (RDoC) \citep{cuthbert2014rdoc} is a modern approach to understand the underlying neurobiology associated with major mental health issues. 
Although a proportion of schizophrenia research has shifted from DSM-ICD to RDoC, we note that the RDoC website states ``RDoC is not meant to serve as a diagnostic guide, nor is it intended to replace current diagnostic systems.'' 
RDoC can be effective supplementary information but the necessary data can be costly to obtain and is therefore less accessible in developing countries, or for patients who must pay to obtain a brain scan but cannot afford to do so. DSM-ICD remains the gold-standard for diagnosis \citep{BARROS2021102039}.
For this reason, we design our diagnostic machine learning system using the traditional symptomatic approach, using DSM-5 symptoms.



\subsection{Artificial Intelligence for Schizophrenia Diagnosis}
DNNs have seen extensive use in the medical domain 
\citep{7801947, zeiser2021deepbatch,rangarajan2021preliminary}. Clinical data is a very popular data source to apply deep learning \citep{8086133} for improved diagnosis \citep{Obermeyer2016}, detection \citep{LAURITSEN2020101820, da2021deepsigns}, and risk prediction \citep{weng2017can}. Kelly \etal \cite{Kelly2019Clinical} identify three challenges for the use of machine learning on clinical data: bias identification, generalisation, and interpretability. 


A vast majority of machine learning applications to mental health focus on the analysis of brain scans \citep{7827160, 8627955, LI2019101696, raza2019diagnosis}.
This is also the case for schizophrenia diagnosis \citep{10.3389/fpsyt.2020.00016, palaniyappan2019effective}.
\cite{Qi2016} apply DNNs to magnetic resonance imaging (MRI) to identify subjects with schizophrenia and schizo-affective disorder.
A layer-wise relevance propagation module was proposed by Yan \etal \citep{Yan2017} to interpret a DNN that discriminated between schizophrenia patients with healthy controls. 
\cite{Kalmady2019} introduce fMRI data that is not corrupted with anti-psychotic medication and develop an ensemble algorithm with multiple parcellations.

Although these results seem promising, the real-world application of machine learning techniques on brain image data for schizophrenia diagnosis remains limited \citep{BARROS2021102039}. There are several problems that are ubiquitous in the domain of machine learning for schizophrenia diagnosis:
\begin{itemize}
    \item There is a large \emph{training-application gap} \citep{wainberg2018deep}, i.e. the trained models accurately describe the source data well, but the models are not shown to be transferable to the real world \citep{tandon2019using, tandon2019machine}, nor are there explanations of how their models could be used in a clinical setting. In contrast, our model is designed to be directly applicable to data that is routinely collected by schizophrenia diagnosticians. Furthermore, we design stress tests to demonstrate that our model would be robust to data from different sources.
    \item Data sets consist of only schizophrenic or healthy individuals but do not contain other unhealthy samples with bipolar disorder or drug overdoses \citep{winterburn2019can}. In practice, a model trained on this data is only able to determine if a patient is healthy or unhealthy, but cannot diagnose schizophrenia. Our data contains patients with multiple causes for the observed symptoms (`bipolar affection disorder', `complex partial seizure', `drug related disorder', etc.) resulting in a more challenging and realistic data set.
    \item Brain image scans in schizophrenia data sets often contain patients who have already started taking anti-psychotic drugs. These drugs have a distinguishable effect on the brain \citep{lesh2015multimodal}, reducing the reliability of the data sets and the results. 
\end{itemize}

\begin{table*}[t]
\centering \caption{Data Set Attributes Relating to DSM-5 Criteria}\label{tab:DSM5}
\scriptsize
\begin{tabular}{p{0.2\linewidth}p{0.3\linewidth}p{0.4\linewidth}}
\hline
DSM-5 Criteria & Attribute(s) & Label Information\\
\hline
Delusions &  `Thought Content' &  Includes persecutory and grandiose delusions\\
Hallucinations & `Thought Perception' & Includes auditory, visual, and tactile hallucination\\
Disorganised speech & `Speech' & Normal, mute, irrelevant or incoherent\\
Grossly disorganised /catatonic behaviour & `Mood', `Attention', `Concentration' & Good, poor, neutral\\
Negative symptoms & `Mental State Examination' & Kempt, unkempt, poor eye contact, restless\\
\hline
\end{tabular}
\end{table*}

\subsection{Interpretable Deep Learning}
Deep learning has high requirements before it can be used in real-world medical applications due to the absolute necessities of trust \citep{trustML}
and interpretability \citep{Vellido2019, rebane2020exploiting}. \emph{Attention} is commonly used to handle these issues \citep{Cheung2018, Park_2018_CVPR}.
The attention modules highlight important structures within data to assist classification performance.
By inspecting these modules, we can better interpret how the neural network comes to a classification decision.
There are various types of attention module, including self-attention \citep{Vaswani2017}, squeeze and excitation \citep{Hu_2018_CVPR} and spatial attention \citep{Wang2017}. 

Multiple works within the medical domain have used attention for interpretability. Choi \etal \citep{Choi2016} propose RETAIN, a two-level neural attention model which processes electronic health records to identify important past visits.
They then use this information to predict the likelihood of heart failure.
Oktay \etal \citep{Oktay18} build Attention U-net to segment organs on an abdominal computed tomography (CT) scan.
This architecture could also be applied for disease diagnosis to segment regions of the image that the network considers important to arrive at a decision.
Paschali \etal \citep{paschali2019deep} equip a DNN with a multiple instance learning branch to output a logit heatmap of the neural activations.

Fused attention mechanisms have shown to improve performance for image processing 
\citep{organisciakmakeup} and natural language processing \citep{duan2018attention}. However, we are unaware of attention fusion for medical diagnostic purposes.

\section{Data Collection}\label{sec:data}
In this section, we motivate the collected data set, give an overview of its characteristics, and detail how it follows current diagnostic guidelines. We also discuss the selected subjects and the pre-processing performed on the completed data set.

\subsection{Data Motivation}
Artificial intelligence for schizophrenia diagnosis has received criticism for not being transferable to the real-world \citep{tandon2019using}. One reason for this is a disconnect between machine learning practitioners and schizophrenia diagnosticians. Machine learning suffers from the ubiquitous practice of competing for the highest scores on benchmark data sets, often giving less regard to real-world applicability. 

\begin{itemize}
    \item The collected data is standardised, based on symptoms from DSM-5 \citep{DSM5}. Attributes are taken directly from case files of patients, so the data will be very similar to that which is already collected in mental health clinics. Moreover, this data does not depend on advanced technology (the EEG attribute could easily be discarded), meaning that models developed from this data are more accessible to developing countries with limited resources. Note that our model requires minimal computational resources.
    \item Studies have shown a link between childhood trauma and schizophrenia in both high income \citep{matheson2013childhood,varese2012childhood} and developing \citep{mall2020relationship} countries. Schizophrenia can be particularly dangerous in poorer regions of Africa, where the likelihood of trauma is greater, the availability of treatment is lower, and there is stigma attached to mental illness \citep{burns2012social}. Most recent works that apply AI to diagnose schizophrenia use images from EEG or fMRI scans \citep{winterburn2019can}. However, these scans are often unavailable in poorer regions and there is no evidence to suggest that models developed in high-income countries can generalise to data from developing countries, either because of hardware differences or genetic differences \citep{gulsuner2020genetics}.
    \item The data allows us to predict a clinician's six-month diagnosis from one observational session. This greatly advances the speed that a potential schizophrenia diagnosis could be identified to help inform the clinician's future care of the patient.
    \item Most research on schizophrenia diagnosis only classify samples with schizophrenia from healthy samples. Mental health clinics rarely face this challenge as patients who do not have schizophrenia will usually have a similar mental health condition. The main difficulty of diagnosis is differentiating schizophrenia from other conditions rather than from healthy samples. Our data set contains samples with a broad variety of health issues to more accurately estimate the difficulty that clinicians face.
\end{itemize}

\subsection{Data Acquisition}\label{sec:acquisition}
The data was collected from a total of 151 subjects, between 2013 and 2018 inclusive, from the Lagos University Teaching Hospital, Lagos, Nigeria.
Our data set consists of psychiatrists' direct observations of their patients; consequently, our model gets the same information that psychiatrists have used to determine their diagnosis.

97 samples are positively diagnosed Schizophrenia patients, and the remaining 54 are other patients with different afflictions.
It is important to note that the negative samples are patients suffering from other related illnesses; they are not healthy individuals as in most schizophrenia diagnosis data sets used for machine learning research.
This makes schizophrenia diagnosis more difficult, but is a significantly more realistic challenge.

The medical records of the patients were obtained from their case files, after obtaining ethical approval from the Lagos University Teaching Hospital Health Ethics Committee.
Each subject signed a patient consent form prior to their mental health assessments; the form permits the use of medical information for the purposes of research and education.
All records were obtained anonymously in line with ethical approval guidelines, to retain the privacy of the subjects.

The full details of the collected attributes can be found in the appendix. Note that although we collect attributes of `Age', `Sex', `Occupation History', and `Marital Status', we do not use them in any of our experiments. RobIn is trained on observed features related to mental health, rather than the general characteristics of the individuals; that is, we train with features that represent an underlying cause of schizophrenia.

With any human-labelled data set, there is a small chance of label error. However, if ML methods can consistently infer a diagnosis from input features, the overall data set can be seen as reliable enough to generalise to new data.

\begin{figure*}[t]
\centering
\includegraphics[width=\linewidth]{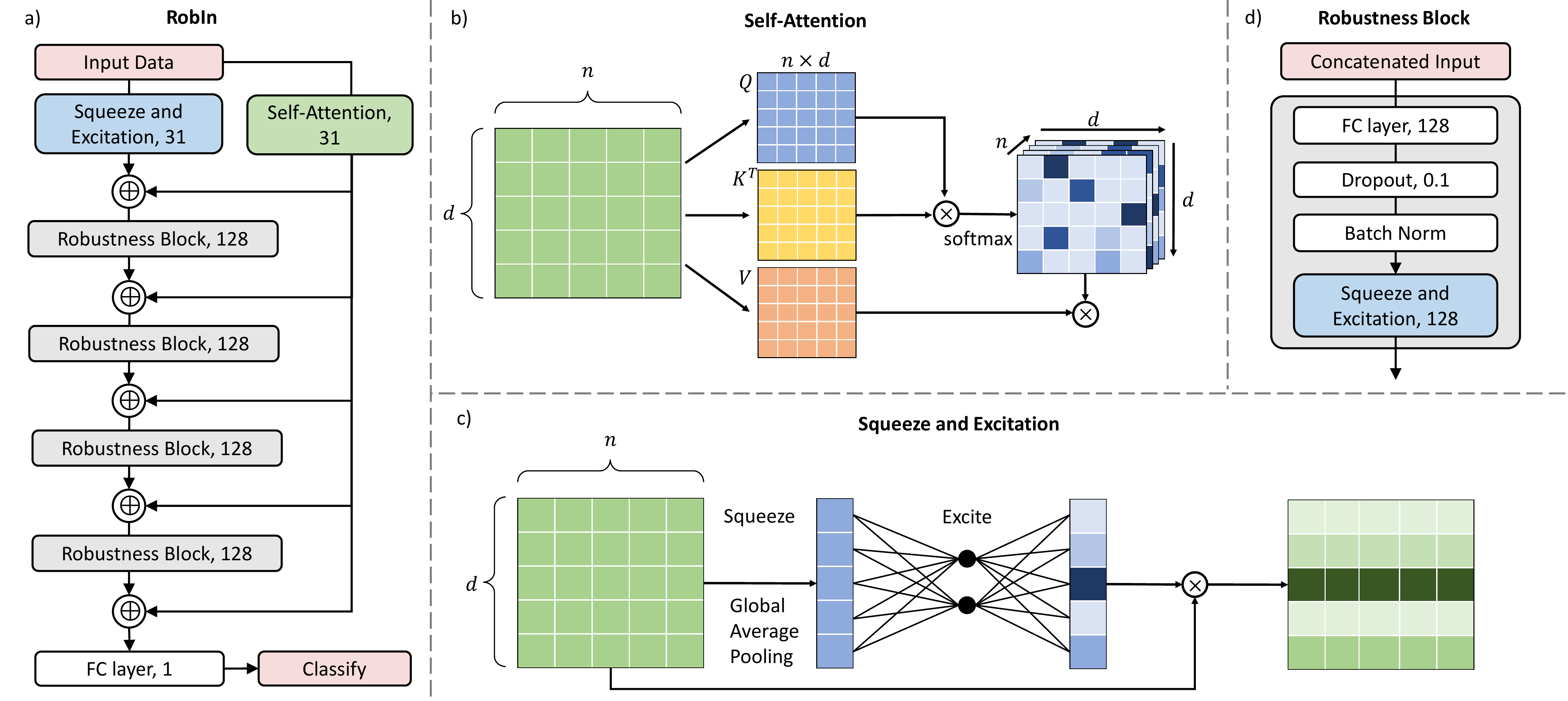}
\caption{a) An overview of the entire network {\color{black} with output dimensions of each layer represented.} Input data goes down the robustness stream and the interpretability stream; b) self-attention mechanism: the input representation is converted into a key, query and value matrix, the cosine distance between each query and each key is found via a matrix multiplication with a higher activation signalling higher alignment between query and key; c) squeeze and excitation: each attribute $i = 1,\ldots, d$ is \emph{squeezed} down to a representative number, a miniature neural network \emph{excites} the squeezed information to evaluate how important each attribute is, then the initial data is multiplied by the importance scores; d) the robustness block we propose in this paper.} \label{fig:SE}
\end{figure*}

\subsection{Data Structure}\label{sec:datastructure}
The data set has 31 raw attributes that are used for learning, and a final diagnosis. 
The attributes collected reflect the criteria used by psychiatrists for Schizophrenia diagnosis, as officially recommended in DSM-5 documentations \citep{DSM5}.
DSM-5 outlines diagnostic criteria A-F, which must be fulfilled for a patient to be diagnosed with schizophrenia.

Criterion A concerns the characteristic symptoms of the condition, and is where DNNs can be effective.
Table \ref{tab:DSM5} presents: the DSM-5 diagnostic criterion A, the attributes selected to reflect criterion A symptoms, and further information on the potential labels of these attributes.

We also collect attributes that commonly occur in other mental health conditions, including depression, bipolar disorder, and schizo-effective disorder. 
We do this because DSM-5 Criteria D-F are exclusions of substance misuse and other disorders such as such as schizo-affective disorder and autism spectrum disorder.


\subsection{Data Pre-processing}
Data pre-processing is an essential step in most machine learning pipelines to improve consistency of the algorithm.
The collected data set contains a mixture of numeric and text data, which is difficult for a deep learning model to process. Categorical data is converted directly into numeric representations via \emph{label encoding}: for every attribute in the data set, each unique value is assigned a representative number. 

Of the selected 31 raw attributes, the majority of samples are missing at least one value. This scenario is common in medical data and is therefore well-studied because, for example, different clinicians record different data. To address this, we set all missing values to $-1$ before performing label-encoding. This method is easiest to apply to the real world where the majority of samples will also have missing values, rather than having to use imputation methods from external data sources that may not match up to the data being processed. 

Experimental results showed that setting missing values to $-1$ resulted in better performance than attempting single-variate or multi-variate imputation techniques. This is likely because this format explicitly informs the system about data absence, allowing the attention mechanism to learn to assign less emphasis.

\section{Deep Neural Networks}\label{sec:DNN}

Deep Neural Networks (DNN) are biologically inspired structures composed of layers of neurons. Every instance from the input data is passed forward through each DNN layer, until it becomes a one-dimensional feature representation of the initial input. This process is repeated many times to improve the quality of the representation.

Let $f^*: D \longrightarrow \{0,1\}^n$ be the function which correctly maps patient data, $\bx_i\in X$, to the true diagnosis, $y_i$,  for all $i=1, \ldots, n$; i.e. 
\begin{equation}
y = f^*(\bx)\fs
\end{equation}
Here, $n$ is the total number of patients, $X\subset \R^{k\times n}$ is collected data with $k$ features of $n$ patients and $\bx_i$ refers to the data of patient $i$.

A DNN learns the function $f: X \longrightarrow \{0,1\}^n$ to get as close to $f^*$ as possible.
For $i = 1, \ldots, n$, we obtain
\begin{equation}
\hat{y_i} = f(\mathbf{x}_i; \theta)\com
\end{equation} 
where $\hat{y_i}$ is the predicted diagnosis of patient $i$, $\mathbf{x}_i \in X$ is the observed data of patient $i$, and $\theta$ are the weights and biases that the DNN learns. The goal is to minimise deviation from the predicted labels $\hat{y}$ to the true labels $y$.

The DNN $f$ is comprised of multiple layers $f_1, \ldots, f_L$ given by:
\begin{alignat}{2}
 \bh_1 &= f_1(\bx; \bW_l, \bb_l) &&= \phi(\bW_0\!^\top \bx 
 + \mathbf{b}_0)\com\\
\bh_l &= f_l(\bh_{l-1}; \bW_l, \bb_l) &&= \phi(\bW_l\!^\top \bh_{l-1} + \mathbf{b}_l)\com
\end{alignat}
where $\phi$ is the non-linear activation function, $\bx$ is the input data, $\bh_l$ is the output of the $l^\mathrm{th}$ hidden layer, $\bW_l$ and $\bb_l$ are the neuron weights and biases of hidden layer $l$ respectively.

We use binary cross-entropy as the loss function:
\begin{equation}
  \mathcal{L}_{\mathrm{BCE}} =  - \sum_{x\in X} (y \log(p(x)) + (1-y)\log(1-p(x))),
\end{equation}
where $y$ is the class and $p$ is the probability assigned by RobIn that an observation $x$ belongs to the class $y=1$.




\section{Interpretable Artificial Neural Network}\label{sec:HUANN}
Despite the excellent performance of deep networks, they suffer from not being interpretable. 
Features at each layer are generated automatically by the network, so it is not clear to humans what these mean. To use machine learning for health applications, it is important to have a degree of understanding of how the algorithm arrives at a conclusion.

We develop a robust, interpretable framework based on \emph{Squeeze and Excitation} and \emph{Self-Attention} to provide insight on the network's decision process. 
From the input layer, all input features get processed by a robustness pathway, governed by squeeze and excitation, and an interpretability pathway, indicated via self-attention.
After each robustness block, we fuse the output vector with the  self-attention, to ensure the interpretable mechanism is influential at all stages of the network. 
Fusing these attention modules allows the network to discover both \emph{global importance} through squeeze and excitation, and \emph{feature interactivity} through self-attention. We are therefore able to interpret the network on multiple levels.
Our overall framework can be found in Figure \ref{fig:SE}.

\subsection{Squeeze and Excitation}\label{sec:SE}
A \emph{Squeeze and Excitation}  unit \citep{Hu_2018_CVPR} is a channel-wise attention mechanism that has recently been proposed for convolutional neural networks (CNN).
It learns the most informative channels at each layer and assigns them greater weight and reduces the weight of less important channels.
We re-design squeeze and excitation for non-sequential data in order to learn the importance of the input channels to help humans understand why a machine has come to a certain conclusion.

First, the information from each attribute is \emph{squeezed} into a representative descriptor via Global Average Pooling. {\color{black} Given the $j^\mathrm{th}$ attribute $a_j = [a_{1j}, \ldots, a_{bj}]$, where $b$ is the number of samples in the batch, we form the attribute descriptor, $c$, via}:
\begin{equation}
c_j = \operatorname{squeeze}(a) := \frac{1}{b} \sum_{i=1}^b a_{ij}\fs 
\end{equation}
These feature descriptors form a vector $\mathbf{z} = [c_1, \ldots, c_d]$ where $d$ is the dimensionality of each sample.

We then \emph{excite} the network to determine which attributes are most important. $\mathbf{z}$ is passed through a fully connected layer, a ReLU, another fully connected layer and finally a sigmoid activation.

Formally, this excitation is written as:
\begin{equation}\label{eq:excite}
\mathbf{s} = \operatorname{excite}(\mathbf{z}) := \sigma\left(\mathbf{W_2} \delta(\mathbf{W_1} \mathbf{z})\right)\com
\end{equation}
where $\mathbf{W_1}, \mathbf{W_2}$ are the parameters of the fully connected layers, $\delta$ is the ReLU function and $\sigma$ is the sigmoid activation function.

\subsection{Self-Attention}
\emph{Self-attention} \citep{Vaswani2017} has transformed the field of natural language processing and is beginning to make similar progress in computer vision \citep{NEURIPS2019_SASA} and graphical data \citep{velickovic2018graph}. We explore its potential for tabular data.

Self-attention learns three new representations for each input feature: a key, query, and value. Give an input feature, $i$: the value vector is a hidden representation of that feature consisting of $d$ attributes; the associated key vector indexes the value vector; and the query vector learns to find keys that are relevant to the value vector. 

Given a query, $q_i$, we calculate attention score $a_{ij}$ of all keys, $k_j$, using the cosine similarity:
\begin{equation}
     a_{ij} = q_i \cdot k_j = ||q_i||\times ||k_j||\cos(\theta)\fs
\end{equation}  
Noting that the dimensionality of all queries and keys remains constant, the more similar $q_i$ and $k_j$ are, the larger $q_i \cdot k_j$ will be. 
A softmax function is applied to polarise the attention weights for each query. Finally, the value matrix, $V$, is then multiplied by these attention scores. This is demonstrated in Figure \ref{fig:SE} b). The entire process can be neatly summarised as
\begin{equation}\label{eqn:SA}
    \operatorname{Attention}(K, Q, V) = \operatorname{softmax}(K^\top Q)V\com
\end{equation}
with $K = \mathbf{W}_k x$, $Q = \mathbf{W}_q x$ and $V = \mathbf{W}_v x$, where $\mathbf{W}_k\com \mathbf{W}_q \com $ and $\mathbf{W}_v$ are learned weights matrices for keys, queries and values, respectively.

To interpret the model, self-attention provides more information than squeeze and excitation. Rather than showing just the raw importance of each feature, self-attention shows us the importance of every feature $j$ to each feature $i$. For example, squeeze and excitation may suggest that a feature $i$ is more important than feature $j$, however it may be the case that $j$ is important if $i$ takes on a particular value (or is missing). Self-attention can handle this more complex circumstance.





\subsection{Robust Interpretable Network}
The architecture is designed with three principles in mind: performance, interpretability, and robustness. We marry the two attention mechanisms described, with squeeze and excitation contributing to performance and robustness, and self-attention mostly benefiting performance and interpretability.

In particular, squeeze and excitation has been widely shown to provide more informative features when used at each successive layer. We follow this standard design for the main branch of our network. Self-attention on the other hand is often used sparingly due to its computational cost being $\mathcal{O}(d^2)$ where $d$ is the number of features. This occurs due to the matrix multiplication in Equation \ref{eqn:SA}. Another issue is that we wish to apply self-attention directly to the data to be able to interpret the final result; however, self-attention applied in early layers often performs an averaging effect in early layers, and only starts to highlight important feature interactions at later layers \citep{ramsauer2020hopfield}. This problem is exacerbated when multiple self-attention modules are used throughout the network.

To solve these problems, we include one self-attention mechanism in our framework, and concatenate its output with the feature representation obtained from each layer.
This means that the self-attention mechanism acts as a direct path from the input data set to the output classification.
This means that the weights of the self-attention mechanism are partially backpropagated directly from the classification layer, rather than the indirect path that is usually seen.
This is important for interpretability because the attention heatmaps have to be powerful enough to contribute to the final classification, otherwise performance would heavily suffer. 
We also find that this self-attention module returns more descriptive, polarised attention heatmaps compared with a traditional implementation, so clinicians can attain more insight from studying them. {\color{black} Each epoch requires 0.012 seconds to train. Total training time lasted 23 seconds.}
Our full RobIn architecture is shown in Figure \ref{fig:SE}.

\section{Evaluation}\label{sec:eval}
\subsection{Evaluation Protocol}
We consider two settings to test machine learning models on our schizophrenia data set:
\begin{enumerate}
    \item 10-fold cross-validation,
    \item A 50/50 train/test split repeated 25 times, to test the ability of all models with less training data, and to have more test data to perturb with the designed stress tests to more reliably evaluate the performance of all models under a different distribution.
\end{enumerate}

We performed hyperparameter tuning on all models with manual search. In all cases, we report the strongest performance that we could obtain. We also report confidence intervals at the 95\% significance level. For fairness, all experiments are performed with the same random seed. The implementation of all frameworks was performed in PyTorch
on an NVIDIA Geforce GTX 2080ti. The full code will be released upon acceptance.

\begin{table*}[t]
\footnotesize
\caption{Comparison with Baseline Machine Learning Techniques for the Diagnosis of Schizophrenia with 90/10 Cross Validation}\label{tab:cv}
\centering
\begin{tabular}{|l|c|c|c|c|c|c|}
\hline
Method & Acc. & F1-score & AuC & Precision & Sensitivity & Specificity \\
\hline
MLP & $91.33 \pm 3.63$ & $92.84 \pm 3.03$ & $91.11 \pm 3.85$ & $89.83 \pm 5.27$ & $96.87 \pm 2.60$ & $85.36 \pm 7.59$\\
DNN & $92.00 \pm 2.70$ & $93.19 \pm 2.46$ & $92.49 \pm 2.85$ & $93.19 \pm 4.04$ & $93.67 \pm 2.86$ & $91.31 \pm 4.94$\\
SVM	& $94.67 \pm 3.14$ & $95.55 \pm 2.71$ & $95.62 \pm 2.52$ & $99.00 \pm 1.62$ & $92.90 \pm 5.00$ & $98.33 \pm 2.71$\\
Tree & $96.00 \pm 3.31$ & $96.84 \pm 2.65$ & $97.15 \pm 2.38$ & $\mathbf{100.00} \pm 0.00$ & $94.29 \pm 4.77$ & $\mathbf{100.00}
\pm 0.00$ \\
\hline
SENN & $96.67 \pm 2.91$ & $97.39 \pm 2.25$ & $97.22 \pm 2.51$ & $98.89 \pm 1.80$ & $96.11 \pm 3.31$ & $98.33 \pm 2.71$ \\
SANN & $96.67 \pm 3.33$ & $97.12 \pm 2.92$ & $96.51 \pm 3.32$ & $96.64 \pm 2.79$ & $97.78 \pm 3.61$ & $95.24 \pm 7.30$ \\
\textbf{RobIn} & $\mathbf{98.00} \pm 2.30$ & $\mathbf{98.56} \pm 1.62$ & $\mathbf{98.33} \pm 3.33$ & $99.00 \pm 1.62$ & $\mathbf{98.33} \pm 2.71$ & $98.33 \pm 5.00$\\
\hline
\end{tabular}
\end{table*}

\begin{table*}[t]
\footnotesize
\caption{Comparison with Baseline Machine Learning Techniques for the Diagnosis of Schizophrenia on a 50/50 Train/Test Split (25 runs)}\label{tab:results}
\centering
\begin{tabular}{|l|c|c|c|c|c|c|}
\hline
Method & Acc. & F1-score & AuC & Precision & Sensitivity & Specificity \\
\hline
SVM	& $83.57 \pm 1.47$ & $ 86.67 \pm 1.28 $ & $83.38 \pm 1.44$ & $89.27 \pm 1.76 $ & $84.78 \pm 2.46$ & $81.97 \pm 2.96 $\\
MLP & $84.43 \pm 1.80$ & $88.24 \pm 1.35$ & $82.14 \pm 2.18$ & $85.04 \pm 2.49$ & $\mathbf{92.22} \pm 1.37$ & $72.07 \pm 4.46$\\
DNN & $84.69 \pm 2.30$ & $88.37 \pm 2.30$ & $82.53 \pm 1.85$ & $85.34 \pm 2.30$ & $92.12 \pm 1.51$ & $72.93 \pm 3.90$\\
Tree & $85.49\pm1.85$ & $88.15 \pm 1.68$ & $85.33 \pm 1.92$ & $ 90.29 \pm	2.17$ & $ 86.71 \pm 2.62$ & $\mathbf{83.95} \pm 3.51$ \\
\hline
SENN & $85.97 \pm 1.69$ & $\mathbf{89.68} \pm 2.32$ & $85.45 \pm 1.99$ & $89.68 \pm 2.32$ & $88.47 \pm 1.88$ & $82.43 \pm 3.94$ \\
SANN & $86.40 \pm 1.82$ & $89.36 \pm 1.47$ & $85.30 \pm 2.35$ & $88.66 \pm 1.47$ & $90.47 \pm 1.60$ & $80.14 \pm 4.02$ \\
\textbf{RobIn} & $\mathbf{86.45} \pm 1.46$ & $89.19 \pm 1.19$ & $\mathbf{86.08} \pm 1.75$ & $\mathbf{90.51} \pm 2.17$ & $88.42 \pm 1.83$ & $83.73 \pm 3.77$\\
\hline
\end{tabular}
\end{table*}

\begin{figure*}
\centering
{\includegraphics[width=0.8\linewidth]{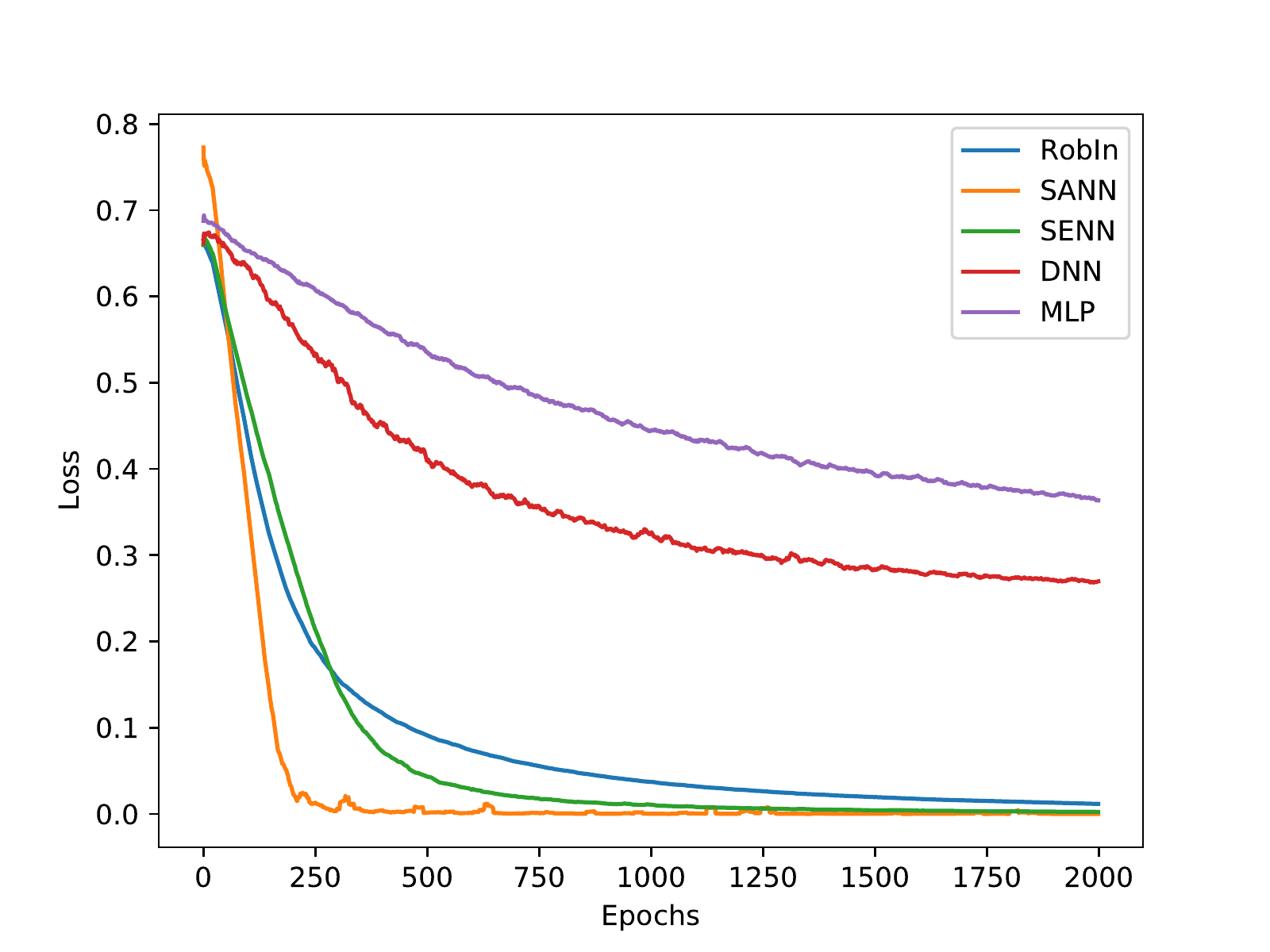}}
\caption{{\color{black}Loss graphs of compared models.} }\label{fig:Loss}
\end{figure*}

\begin{table*}[t]
\footnotesize
\caption{{\color{black} Comparison with State-of-the-Art Methods using Brain Imaging Data {\footnotesize SZ = Schizophrenia, BD = Bipolar Disorder, HC = Healthy Controls, OD = Other Diseases}}}\label{tab:brain_img}
\centering
\begin{tabular}{|l|cccc|}
\hline
Reference & Data Type  & Samples & Classification Type & Accuracy \\
\hline
\cite{Liautoencoder} & fMRI & $183$ & SZ v HC & $80.5$ \\
\cite{Niuaugmentation}	& fMRI & $82$ &  SZ v HC & $90.8$ \\
\cite{palaniyappan2019effective} & fMRI & $57$ &  SZ v BD & $76.3$ \\
\cite{Salvador} & fMRI+sMRI & $211$ & SZ v HC  & $84.0$ \\
\cite{10.3389/fpsyt.2020.00016} & sMRI & 926 & SZ v HC & $88.6$ \\
\cite{Phangmulti} & EEG & $84$ & SZ v HC & $91.7$ \\
\cite{Calhaspairwise} & EEG & $84$ & SZ v HC & $95.0$ \\
\hline
\textbf{RobIn} & Clinical & $151$ & SZ v OD & $98.0$\\
\hline

\end{tabular}
\end{table*}

We compare three attention models that we propose: our \textbf{Rob}ust \textbf{In}terpretable Deep Network, {\bf RobIn}, a squeeze and excitation network with four-layers, {\bf SENN}, and a  self-attention network containing four consecutive self-attention layers (i.e. no fully-connected layers), {\bf SANN}. These models are compared with standard approaches: a decision tree, {\bf tree}, a support vector machine, {\bf SVM}, a neural network with one layer, {\bf MLP}, and a deep neural network consisting of four layers, {\bf DNN}. Note that we can consider the comparison between RobIn, SENN, DNN, and MLP an ablation study where we remove the self-attention mechanism, the squeeze and excitation layers and the additional hidden layers respectively.

We report the metrics: `Accuracy', `Precision', `Sensitivity', `Specificity', `F1 Score' and `AuC', where 
\begin{align}
\mathrm{Accuracy} &= \tfrac{\TP + \TN}{\TP + \TN + \FP + \FN}\com\\
\mathrm{Precision} &= \tfrac{\TP}{\TP + \FP}\com \\
\mathrm{Sensitivity} &= \tfrac{\TP}{\TP + \FN}\com \\
\mathrm{Specificity} &= \tfrac{\TN}{\TN + \FP}\com
\end{align}
are calculated via true positives (TP), true negatives (TN), false positives (FP) and false negatives (FN). From these measures we obtain the F1 Score and AuC: 
\begin{equation}
\mathrm{F1} = 2\cdot\frac{\PPV \cdot \TPR}{\PPV + \TPR}\fs
\end{equation}

The AuC is the area under a receiver operating characteristic (ROC) curve, plot on the unit square with axes sensitivity and (1-specificity). An AuC of 0.5 indicates random guessing.

\begin{figure}
\centering
{\includegraphics[width=\linewidth]{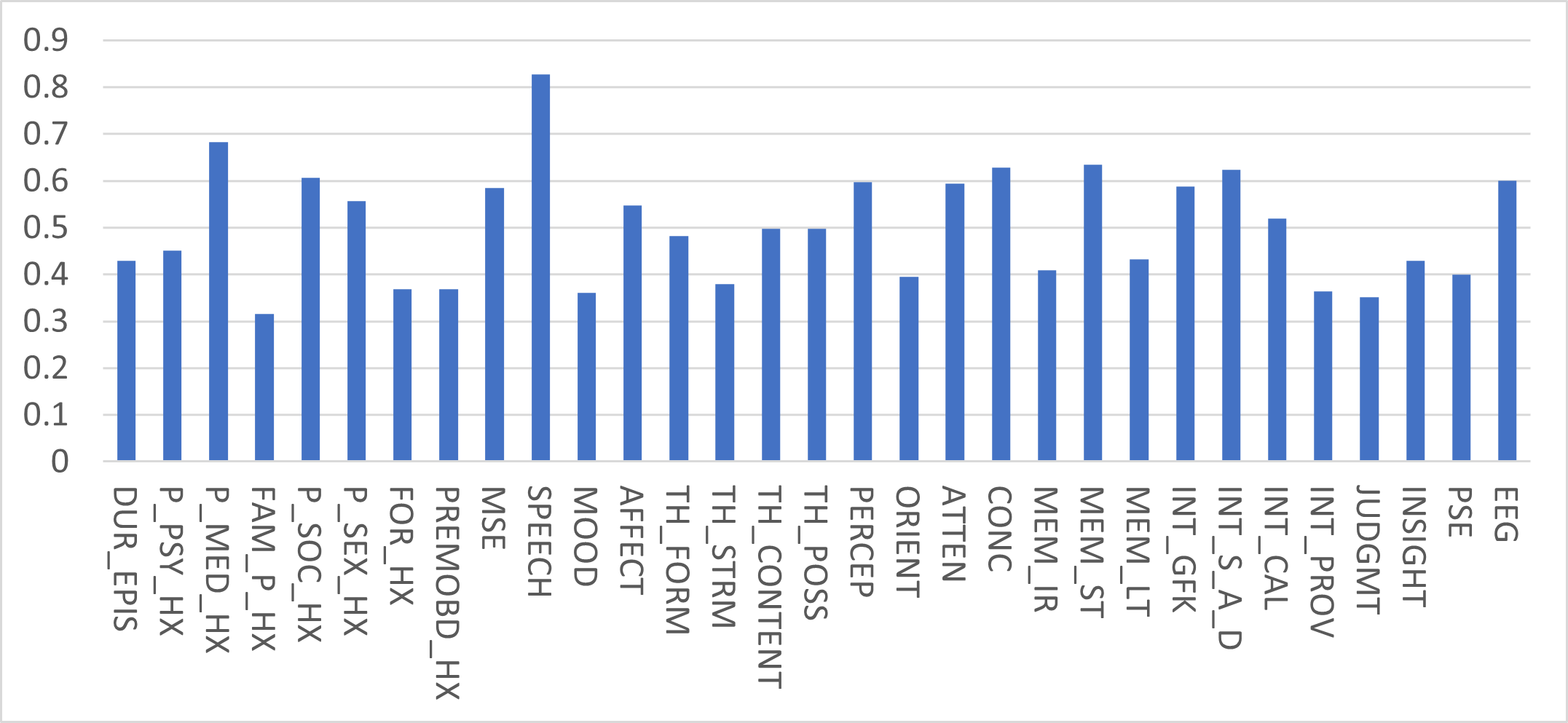}}
\caption{Global Feature Importance of RobIn: as expected, no features are entirely discarded, but certain features such as speech, past medical history and concentration are thought to be important. }\label{fig:InterpretSE}
\end{figure}

\begin{figure*}
\centering
{\includegraphics[width=\linewidth]{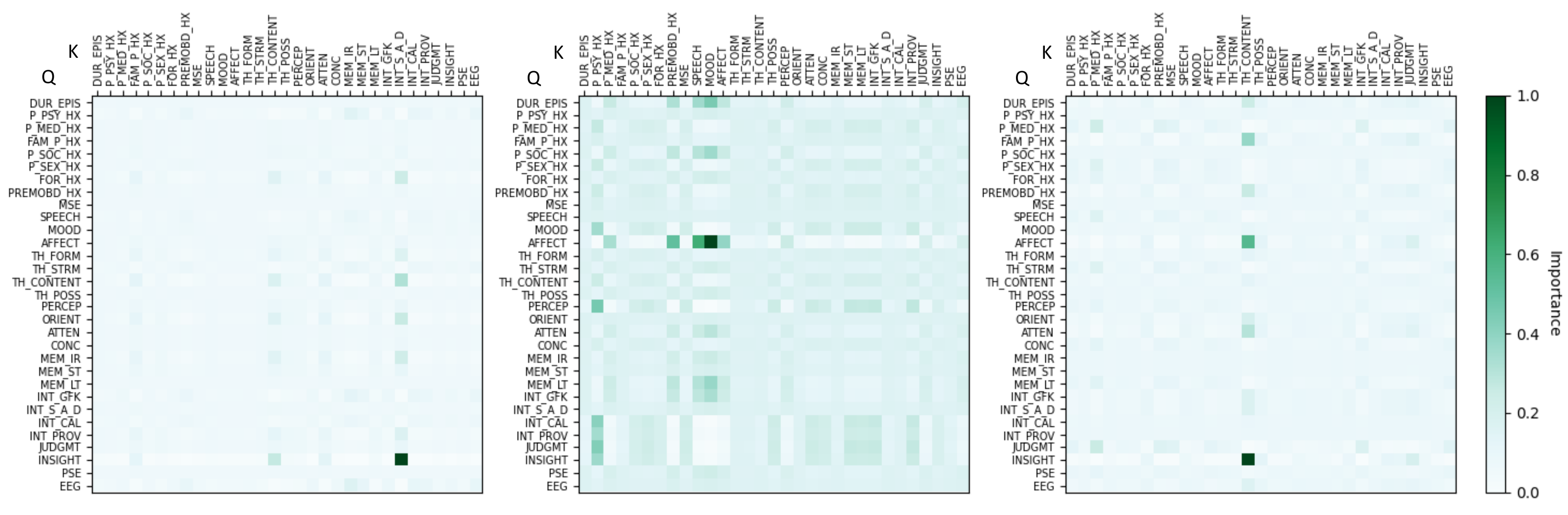}}
\caption{Heatmaps generated by the self-attention module. 
Darker squares indicate higher importance assigned by the queries (rows) to the keys (columns). }\label{fig:InterpretSA}
\end{figure*}
\subsection{Model Performance}
The effectiveness of the compared models on the 10-fold cross-validation setting is evaluated in Table \ref{tab:cv}. RobIn attains the best results across the most descriptive measures: accuracy, F1-score and AuC. RobIn outperforms MLP and DNN to a statistically significant level on all three measures, and SVM on Accuracy and F1-score. Given that the decision tree also obtains a good performance, it is very difficult to get a statistically significant result. Nonetheless, we get a statistically significant improvement on F1-score at a 95\% significance level, and get $p$-values of 0.086 and 0.121 for accuracy and AuC, respectively.

The effectiveness of the compared models over 25 runs on the 50/50 split is evaluated in Table \ref{tab:results}. These results are shown as a pre-cursor to the robustness tests in Section \ref{sec:Stress}.
With a much smaller training set, we expect the traditional models to outperform the deep models. However, with the same random seed, RobIn outperforms all other models on Accuracy and AuC. SENN achieves a higher F1 score than RobIn; however, it does so with a very high standard deviation. Although our results are not statistically significantly better than the decision tree, Section \ref{sec:Stress} shows that the scores from the decision tree and SANN are unreliable because they over-predict a positive diagnosis because the data set has more positive samples than negative. This means that the performance of these methods has been artificially inflated. Although they are capable of predicting from an i.i.d. test set, they cannot be trusted in a clinical setting.

{\color{black} We also provide loss curves of compared models in Figure \ref{fig:Loss}. MLP and DNN are not able to converge to a low loss, indicating that they underfit the training data. On the other hand, SANN reduces towards zero very sharply, but then appears unstable. This indicates that training data is being memorised. RobIn and SENN appear to converge in an appropriate fashion, with RobIn having a shallower slope after 300 epochs, indicating it is least likely to be overfitting.}

\subsection{Comparison with Brain-Imaging Research}

{\color{black} To place our work into context of the ongoing research of using artificial intelligence to diagnose schizophrenia, in Table \ref{tab:brain_img}, we compare with a selection of state-of-the-art works that were conducted using brain imaging data. For a comprehensive study, see \citep{cortes2021going}. 

Most data sets are relatively small, with the second highest number of samples being 211. The exception is \citep{10.3389/fpsyt.2020.00016}, who combine different publicly available data sets. To release the full potential of modern machine learning algorithms, large amounts of data are required. However, brain imaging data is prohibitive to obtain. We believe that out work establishes a proof-of-concept for the use of clinical data to diagnose schizophrenia, which should be easier to obtain and process, allowing for significantly larger data sets to be constructed in the future.

In terms of performance, RobIn applied to clinical data exceeds the performance of all image classifiers. Most brain-imaging methods are only required to discriminate between patients with schizophrenia and healthy controls, whereas we have a more difficult test setting as our control group consists of patients with other mental illnesses. This demonstrates the potential of involving clinical data within diagnostic systems, either alone or in combination with brain-imaging techniques.}

\subsection{Interpretability}
\subsubsection{Global Attribute Importance}

An example of the global attribute importance is shown in Figure \ref{fig:InterpretSE}. At this stage of the network, the squeeze and excitation mechanism needs to consider all attributes, and it therefore does not give any feature a particularly low score. However, we can see that the model determines speech ability to be very important in finding a diagnosis, while other attributes like past medical history, concentration, short-term memory, and EEG scan are also deemed important. 

These scores remain identical for test data that is considered. This allows our system to remain robust to different stress tests that we conduct, which indicates that this global attribute importance is essential for the model to generalise to real-world data.

\begin{figure*}[t]
\centering

\includegraphics[width=\linewidth]{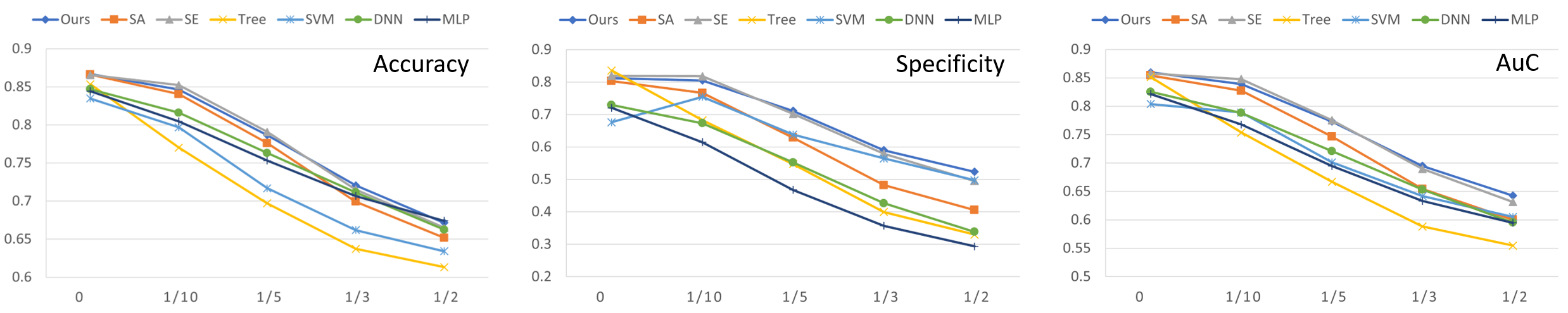}
\caption{
Robustness comparisons with the addition of noise, $X \sim \mathcal{N}(\mu=0, \sigma^2)$, where the $x$-axis is $\sigma^2$. }\label{fig:Noise}
\end{figure*}

\begin{figure*}[t]
\centering

\includegraphics[width=\linewidth]{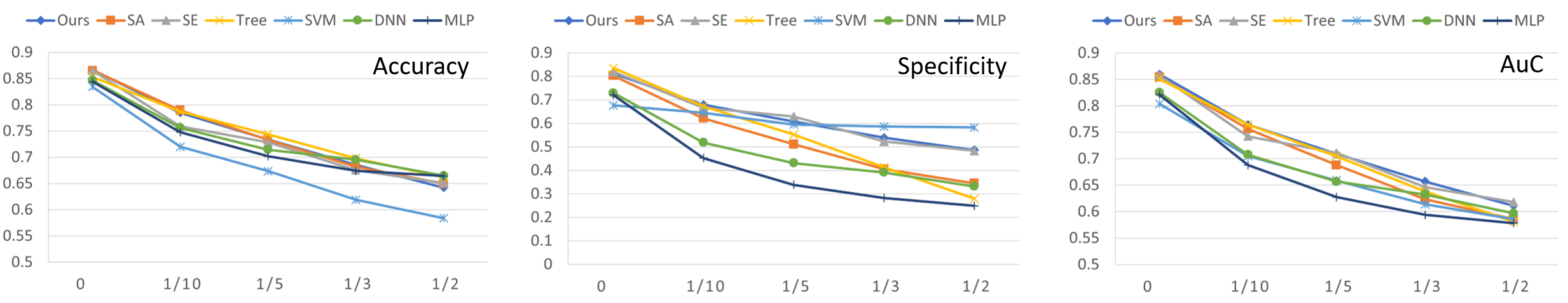}
\caption{
Robustness comparisons with the removal of data points where the $x$-axis signifies the fraction of values that were removed from the test data.  }\label{fig:MissingValues}
\end{figure*}

\begin{figure*}[t]
\centering
\includegraphics[width=\linewidth]{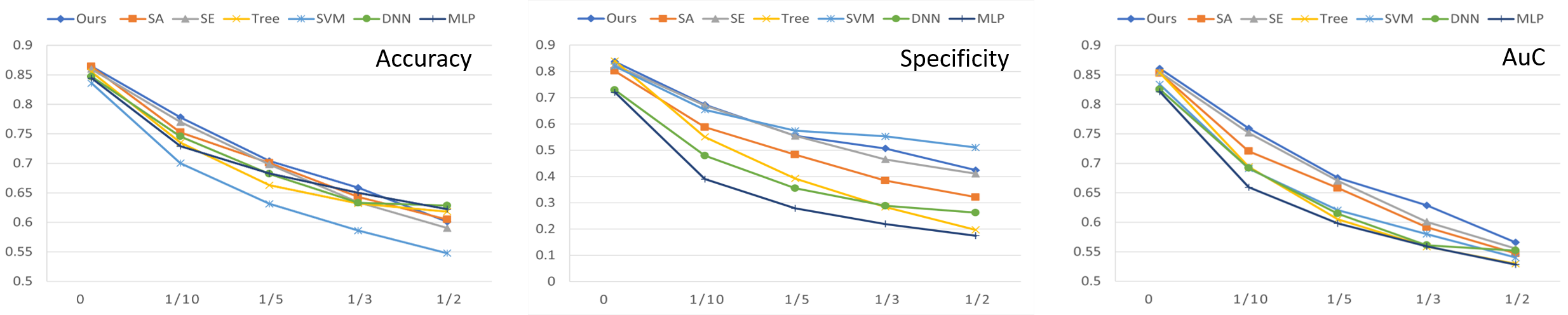}
\caption{
Robustness comparisons with the addition of noise and the removal of data points. The $x$-axis signifies the variance of the normal curve from which the noise was sampled and then added to the test data, and the fraction of values that were removed from the test data.}\label{fig:Both}
\end{figure*}

\subsubsection{Attribute Interactivity}
In Figure \ref{fig:InterpretSA}, we see examples of the attention activations of the self-attention mechanism. These heatmaps visualise the features that are deemed important to arrive at a diagnosis. In particular, we can observe that a model often finds that the insight of a patient is a useful query.

Note that because of the design of RobIn, more polarised activations are obtained because we connect the first layer with the final layer. In self-attention frameworks, the first layer activations are usually averaged while final layer activations are more polarised. However, this is not beneficial for interpreting feature importance because by the time polarised heatmaps are obtained, features have already undergone several layers of abstraction.


\subsection{Stress Tests}\label{sec:Stress}
DNNs are difficult to implement in the real world because, even though they perform well on unseen samples from the same data set, they struggle to generalise to unseen samples from a different source \citep{d2020underspecification}. This is because the test set is independent and identically distributed (iid) to training set, but real-world data rarely has an identical distribution to small-sample collected data sets. We design three stress tests to randomly alter the distribution of the test set at varying strengths to evaluate the robustness of all models:
\begin{enumerate}
    \item {\bf Noise}: We sample noise $X \sim \mathcal{N}(\mu=0, \sigma^2)$ and add this noise distribution to the test data to simulate different ways that a clinician might record a patient observation
    \item {\bf Data Erasing}: We randomly eliminate a fraction $1/m$ of the test data to simulate different ways that missing values could impact performance
    \item {\bf Noise + Data Erasing}: We combine 1) and 2) to provide a challenging, realistic test of model robustness
\end{enumerate}

All models are evaluated on the original 50/50 train/test split with the same random seed, with the test set altered via the aforementioned perturbations. For each stress-test, we evaluate $\sigma^2, 1/m$ at values $\{ 1/10, 1/5, 1/3, 1/2 \}$.
Note that all perturbations are applied after the initial training data has been normalised between $0$ and $1$.

The performances under stress tests are presented in Figures \ref{fig:Noise} - \ref{fig:Both}. We found that accuracy and F1-score were poor indicators of model performance under these stress-tests because the data set is skewed with more positive samples than negative ones. 
As data became more noisy, models would often over-predict a positive diagnosis to obtain the highest accuracy. This can be seen from the specificity graphs, where the decision tree and the neural networks without squeeze and excitation modules all drop below 0.35. This indicates that models have learnt to make decisions based on the data set characteristics, rather than the features provided.
We report the accuracy and specificity to present this phenomenon, but base our robustness evaluation on the AuC - the only reliable evaluation metric because it is a function of specificity. 

The figures show that the decision tree and SANN are not deployable because they show a tendency to over-predict a positive diagnosis. Those models are not suitable for real-world deployment because, counter to our data set, the proportion of people who actually have schizophrenia is very low.

Across all three stress tests, RobIn and SENN suffer least from perturbations of the test data, indicating that they are most suitable for deployment in the real world. This gives credence to our claim that the squeeze and excitation mechanism improves model robustness.

\subsubsection{Additional Noise} As seen in Figure \ref{fig:Noise}, when noise $X \sim \mathcal{N}(\mu=0, 1/2)$ is added, the decision tree does barely better than random guessing with an AuC of 0.555, whereas RobIn maintains an AuC of 0.643 which is very impressive given the severity of the perturbation. Moreover, when $X \sim \mathcal{N}(\mu=0, 1/5)$ perturbs the data, the AuC of RobIn falls only by 0.088, in contrast to the 0.186 drop by the decision tree. In the realistic scenario when $\sigma^2 = 1/10$, all methods using attention see very little performance loss, whereas again the decision tree performs worst with a loss of $0.099$, more than RobIn suffers with $\sigma^2$ is twice as large.

\subsubsection{Removed Values} In Figure \ref{fig:MissingValues}, the removal of values appears to be a much more difficult task, as shown by the sharp drop in AuC even when only $1/10$ values are removed. RobIn and SENN maintain the highest AuCs throughout the stress levels. All models struggle with half of the values removed and are close to random guessing. At the $1/3$ stress level, RobIn has 0.011 greater AuC than the next best, SENN. It is worth noting that the specificity of the SVM remains fairly constant as more values are removed; however, it has by far the lowest accuracy at all stress levels and only consistently outperforms MLP on AuC, despite the high specificity. It over-predicts a negative diagnosis, resulting in a high number of true negatives and a low number of false positives. 

\begin{figure}[t]
\centering
\begin{tabular}{cc}
{\includegraphics[width=0.4\linewidth]{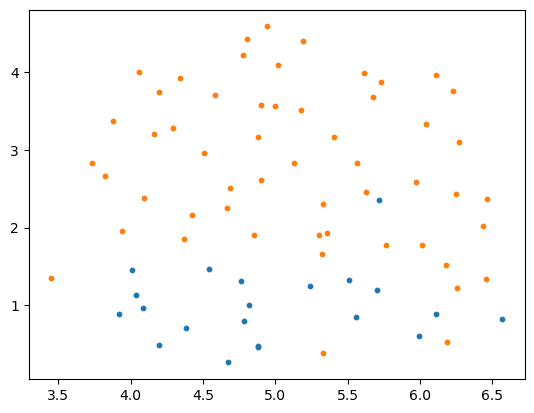}}&
{\includegraphics[width=0.4\linewidth]{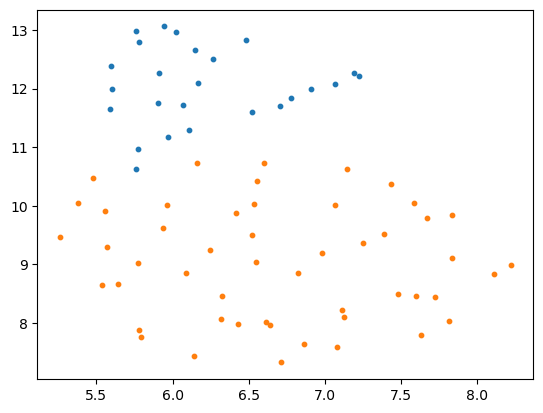}}\\
DNN & SENN\\
{\includegraphics[width=0.4\linewidth]{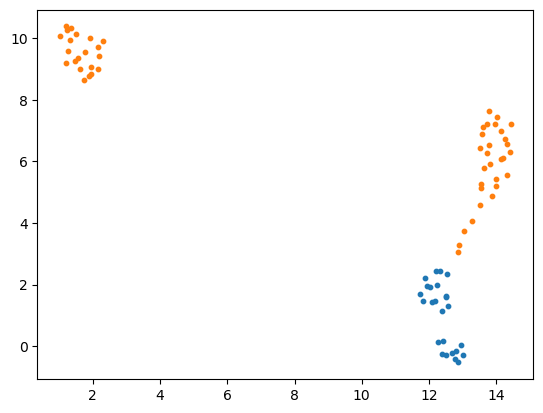}}&
{\includegraphics[width=0.4\linewidth]{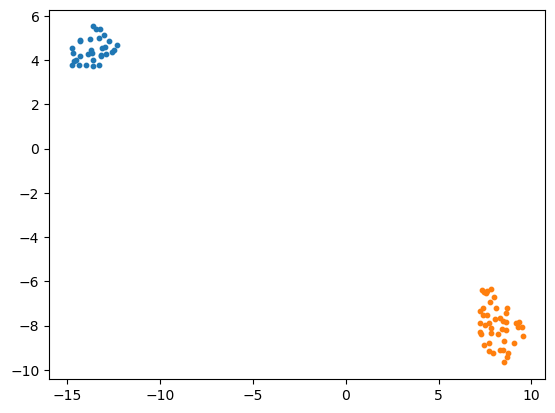}}\\
 SANN & RobIn
\end{tabular}\caption{
UMAP results across tested neural network models. Blue dots are negative samples, orange dots are positive samples.}\
\label{umap}
\end{figure}

\subsubsection{Combination} The combination of the previous two stress tests in Figure \ref{fig:Both} is the most realistic test, where noise simulates difference of opinion between clinicians and the data point removal simulates values that are often missing. RobIn outperforms all non-attention based methods on AuC at $1/10, 1/5,$ and $1/3$ stress levels, with a 95\% confidence interval. Furthermore, in the $1/3$ level, RobIn attains a $0.629 \pm 0.0257$ AuC, a statistically significant improvement on the next best method, SENN, which attains 0.601 AuC. RobIn even maintains the highest accuracy until the $1/3$ level, despite the propensity of other methods to cheat. It is clear that RobIn is the most robust method and most suitable for real-world deployment.

\subsubsection{Discussion of Stress Tests}
{\color{black} The weakness of traditional models at handling these stress tests demonstrates the necessity of conducting evaluation to test robustness of models. Machine learning models often overfit on the data distribution they are trained on, and therefore struggle to generalise to unseen distributions, particularly when training data is small. Conducting these stress tests has highlighted significant flaws in models such as MLPs, SVMs and decision trees, despite the fact they appear to work well on the uncorrupted data.

Furthermore, stress tests appear to be a strong method to evaluate a model's ability to reason with imbalanced data. As data perturbation gets stronger, models that are less robust will classify samples into the majority class, whereas more robust models will continue to reason with the corrupted data. This indicates that more robust models have a more sophisticated ability to retrieve correlations between subsets of features and outputs, so are not overly reliant on individual features that can become corrupted. 
} 

\subsection{Feature Visualisation} 
In Figure \ref{umap}, we use UMAP \citep{mcinnes2020umap} to visualise the 2-dimensional manifold of feature representations of different neural networks.

In the top left, the output of the DNN is disorganised and there is a lot of crossover between the two classes. SENN has a clearer boundary between classes, but there is only a small distance between the two classes; many samples being nearer to samples from the opposite class than their own. In the bottom left, SANN does a better job of clustering samples but there are a subset of positive samples very close to the negative samples. This shows that the model is unsure about a vast percentage of positive samples. 

On the other hand, RobIn does the best job of categorically separating positive and negative samples, with a large distance between the two clusters that are formed. This highly organised output manifold demonstrates the ability of RobIn to categorise data, while also demonstrating the benefits of combining the two attention mechanisms. Squeeze and excitation forms a strong decision boundary whilst self-attention contributes to clustering similar data.

\section{Conclusion and Discussions}\label{sec:conclusion}
Applications of machine learning to schizophrenia diagnosis have struggled to bridge the gap between theory and application.
To combat this, we have collected a data set that complies with current clinical guidelines outlined by 
DSM-5. This data will be of a similar format to records that are already kept by clinicians for patient monitoring.

We have developed a robust, interpretable network that outperforms other machine learning methods for the diagnosis of schizophrenia. Our network contains a squeeze and excitation mechanism that works on a global scale to give an overview of feature importance and a self-attention mechanism that works on a local level to assess feature interactivity. We have also comprehensively tested all methods with stress tests to evaluate their potential to generalise to new data sources. Our proposed method is the most robust of all compared models on all three stress tests. The evidence suggests that this is because of the static weights learnt by the squeeze and excitation mechanism. Therefore, we conclude that self-attention contributes to model interpretability and squeeze and excitation contributes to model robustness, whilst both contribute to overall performance.

{\color{black} For future work, we wish to collect a larger data set of clinical data from several different test sites to further evaluate the performance of RobIn, so that RobIn or similar frameworks could be used in real-world clinical decision making. There are two streams of work that we are particularly interested in: a) \textit{combining clinical and brain-imaging features} - we have demonstrated that deep learning can use clinical data to diagnose schizophrenia, we want to fuse clinical data with associated image data to allow the model to have all of the available information when making a decision, and to use cross-attention to link clinical features with highlighted areas on the fMRI scan; b) \textit{attentional machine feedback for clinicians} - we wish to analyse high-attention features that RobIn finds when training on a larger data set to identify particularly influential features. Current works that utilise brain-imaging aim to identify regions of importance to help improve clinician's ability of diagnosing from brain images themselves. In a similar way, we wish to identify behavioural features that clinicians should pay special focus to when interacting with a client for the first time.}

\section*{Acknowledgement}
The project was supported in part by the Royal Society (Ref: IES\textbackslash R2\textbackslash 181024). 
Declarations of interest: none.

\bibliography{sample}

\appendix

\begin{table}
\caption{Representation of Features, Descriptions and Values}
\scriptsize
\begin{tabular}{lp{0.4\linewidth} p{0.4\linewidth}}
FEATURE & DESCRIPTION & VALUES \\
\hline
AGE & Age of patient & Age e.g. 23, 34 \\
SEX & Sex of patient & Male, female \\
OCCUP\_HX & Occupation history & Unemployed, occupation \\
MAR\_STA & Marital status & Married, single, divorced, widow \\
DUR\_EPIS & Episode duration (length of time the patient has suffered the symptoms) & Time in months \\
P\_PSY\_HX & Past psychiatric history & No, yes (e.g. rape, mental illness, etc.) \\
P\_MED\_HX & Past medical history & No, disease suffered in past (e.g. diabetes) \\
FAM\_P\_HX & Family psychiatric history & Yes, no \\
P\_SOC\_HX & Past social history & Yes, no \\
P\_SEX\_HX & Past sexual history & Normal, experience (e.g. masturbate, gonorrhea, etc.) \\
FOR\_HX & Forensic history & Yes, no \\
PREMOB\_HX & Pre-morbid history & Normal, introvert, extrovert, melancholic \\
MSE & Mental state examination & Kempt, unkempt, poor eye contact, restless \\
SPEECH & Speech status & Normal, reduced volume, mute, slurred, decreased tone, irrelevant, incoherent \\
MOOD & Mood of the patient at the time of report & Euphoric, neutral, happy, relaxed, fine/ok, worried, sad, irritable \\
AFFECT & Affect of the patient at time of report & Depressed, reactive, blunt, restricted, congruent, abnormal \\
TH\_FORM & Thought form at time of report & Logical, abnormal \\
TH\_STRM & Thought stream at time of report & Reduced, normal, increased \\
TH\_CONTENT & Thought content at time of report & Persecutory delusion, auditory hallucination, normal, obsession, grandiose delusion, disorder \\
TH\_POSSESSION & Thought possession at time of report & Impaired, normal \\
PERCEP & Perception at time of report & No, auditory hallucination, visual hallucination, tactile hallucination, olfactory hallucination, preoccupation \\
ORIENT & Time, place and position orientation at time of report & Oriented in TPP, no \\
ATTEN & Attention status at time of report & Rousable, poor \\
CONC & Concentration status at time of report & Good, reduced, poor \\
MEM\_IR & Immediate recall status at time report & Good, fair, poor \\
MEM\_ST & Short-term memory status at time report & Good, fair, poor \\
MEM\_LT & Long-term memory status at time report & Good, fair, poor \\
INT\_GFK & Intelligence test of general fund of knowledge & Good, fair, poor \\
INT\_S\_A\_D & Intelligence test of similarity and difference & Good, fair, poor \\
INT\_CAL & Intelligence test of arithmetic & Good, fair, poor \\
INT\_PROV & Intelligence test of proverbs & Good, fair, poor \\
JUDGMT & Intellectual judgement status at time of report & Good,  poor \\
INSIGHT & Insight status at time of report & Good, partial ,poor \\
PSE & Physical state examination status & Good, normal, pale \\
EEG & Electroencephalogram (to exclude brain tumour possibility status) & Normal, altered
\end{tabular}
\end{table}

\end{document}